\pdfoutput=1

\documentclass[11pt]{article}

\usepackage[utf8]{inputenc} 
\usepackage[T1]{fontenc}    
\usepackage{hyperref}       
\usepackage{url}            
\usepackage{booktabs}       
\usepackage{amsfonts}       
\usepackage{amsmath}        
\usepackage{nicefrac}       
\usepackage{microtype}      
\usepackage[table]{xcolor}         
\usepackage{multirow}
\usepackage{tcolorbox}

\usepackage{graphicx}
\usepackage{longtable}
\usepackage{caption}
\usepackage{geometry}
\usepackage{adjustbox}
\usepackage{float}      

\usepackage{colortbl}  
\usepackage{array}     
\usepackage{amsmath}   
\usepackage{booktabs}  
\usepackage{makecell}  
\usepackage{wrapfig}   

\usepackage{graphicx}
\usepackage{subfigure}
\usepackage{url}

\usepackage{bm}

\usepackage{acl}

\usepackage{times}
\usepackage{latexsym}

\usepackage[T1]{fontenc}

\usepackage[utf8]{inputenc}

\usepackage{microtype}

\usepackage{inconsolata}

\usepackage{graphicx}

\title{Beyond the Trade-off: Self-Supervised Reinforcement Learning for Reasoning Models' Instruction Following}

\author{
 \textbf{Qingyu Ren\textsuperscript{1}\thanks{\ Equal contribution.}},
 \textbf{Qianyu He\textsuperscript{1}\footnotemark[1]},
 \textbf{Bowei Zhang\textsuperscript{2}}, \textbf{Jie Zeng\textsuperscript{1}},
 \textbf{Jiaqing Liang\textsuperscript{2}\thanks{\ Corresponding author.}},
 \textbf{Yanghua Xiao\textsuperscript{1}\footnotemark[2]}\\\textbf{Weikang Zhou\textsuperscript{3}, Zeye Sun\textsuperscript{3}, Fei Yu\textsuperscript{3}}\
\\
    \textsuperscript{\rm 1}Shanghai Key Laboratory of Data Science, College of Computer Science and Artificial Intelligence,\\ Fudan University,
    \textsuperscript{\rm 2}School of Data Science, Fudan University,
     \textsuperscript{\rm 3}Ant Group\\
     \{qyren24,qyhe21,bwzhang24, jzeng23\}@m.fudan.edu.cn, \{liangjiaqing, shawyh\}@fudan.edu.cn
}

\begin{document}
\maketitle
\begin{abstract}
Reasoning models excel in complex problem solving but exhibit a concerning trade-off between reasoning capabilities and instruction following abilities.
Existing approaches for improving instruction following rely on stronger external models, creating methodological bottlenecks and practical limitations including increased costs and accessibility constraints.
We propose a self-supervised RL framework that leverages reasoning models' own internal signals to improve instruction following capabilities without external supervision.
Extensive experiments demonstrate that our framework significantly improves instruction following capabilities while maintaining reasoning performance, offering a scalable and cost-effective approach to enhance instruction following in reasoning models.
The data and code are publicly
available at \href{https://github.com/Rainier-rq/verl-if}{https://github.com/Rainier-rq/verl-if}.
\end{abstract}

\section{Introduction}


\begin{figure}[t] 
    \centering
            \includegraphics[width=0.4\textwidth]{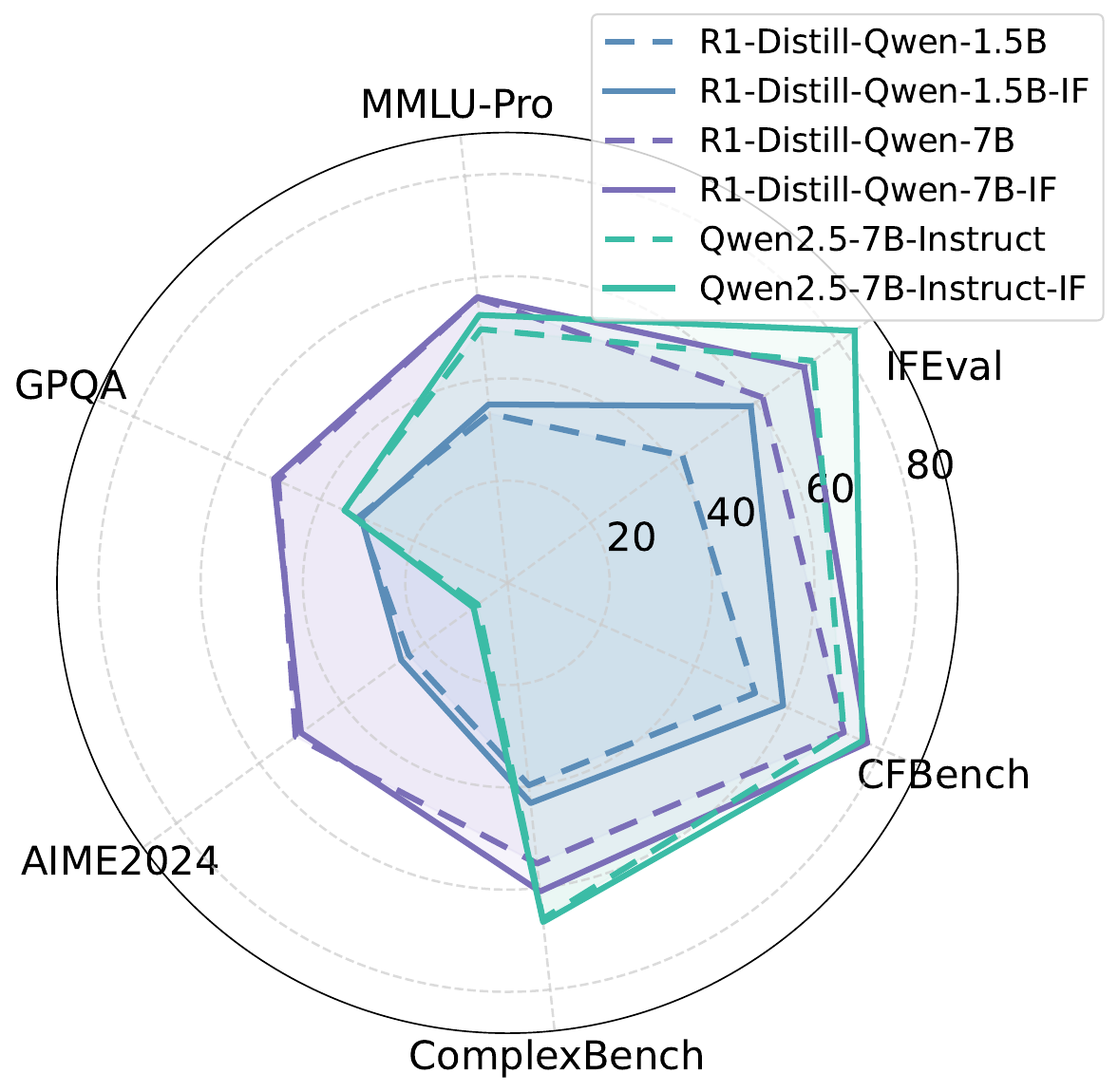}
    \caption{Performance comparison of distilled, instruct models and their variants optimized by our method (-IF) on reasoning tasks (AIME2024, GPQA, MMLU-Pro) and instruction following tasks (IFEval, CFBench, ComplexBench).}
    \label{fig:motivation}
\end{figure}
\begin{figure*}[t] 
    \centering
            \includegraphics[width=0.9\textwidth]{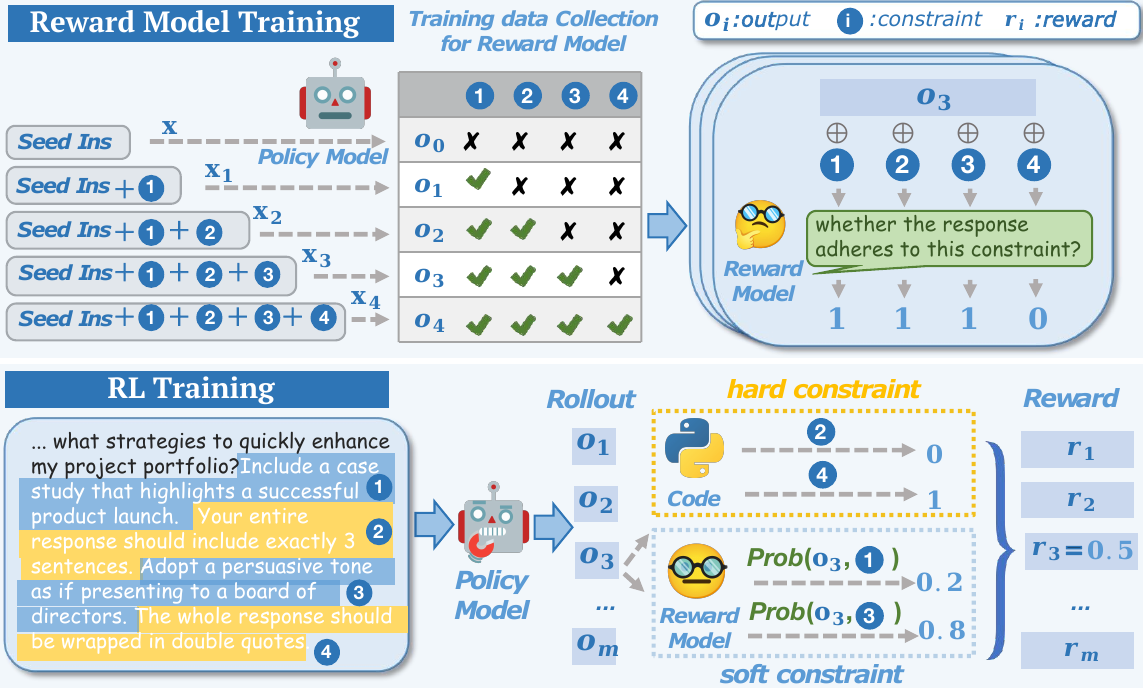}
    \caption{Overview of our RL framework. We first self-supervisedly construct reward model training data through curriculum decomposition for constraint-wise binary classification. Then, we efficiently train the policy model with both hard constraints via rule-based verification and soft constraints via the reward model.}
    \label{fig:framework}
\end{figure*}
Reasoning models have excelled in various reasoning domains~\cite{o1, guo2025deepseek, seed2025seed1}.
Their instruction following capabilities (i.e., the ability to follow multiple constraints simultaneously) are crucial to ensure practical use in real-world applications.
On one hand, real-world conversations with human users often contain multiple constraints in the instructions~\cite{deshpande2025multichallenge}.
On the other hand, reliable instruction following is essential for reasoning models to excel in complex agentic tasks~\cite{qi2025agentif}.

However, reasoning models exhibit a concerning trade-off between reasoning capabilities and instruction following abilities.
Fig.~\ref{fig:motivation} illustrates this phenomenon.
Current approaches show a clear bias: instruction-tuned models excel in instruction following while inferior in reasoning capabilities~\cite{team2024qwen2}, while reasoning models prioritize reasoning performance but underperform in complex instruction following tasks~\cite{guo2025deepseek}.
This trade-off poses challenges for reasoning models in real-world applications that require both capabilities.

Existing approaches for improving instruction following abilities have a key limitation: they rely on stronger external models.
Some works distill training data from stronger models for supervised fine-tuning~\cite{sun2024conifer} or collect pairwise preference data for direct preference optimization~\cite{he2024complex}.
Others adopt reinforcement learning (RL), but require stronger models as reward models or for distilling reward model training data~\cite{qin2025incentivizing}.
This dependency creates several problems.
Methodologically, the teaching model's capabilities limit the student model's potential improvement~\cite{xu2024stronger}.
Practically, this approach introduces increased accessibility constraints, as stronger models are often proprietary or computationally expensive~\cite{sun2024conifer}.

Hence, we leverage the reasoning model's own internal signals to improve its instruction following capabilities.
To achieve this goal, we choose RL as our training paradigm for several reasons.
First, RL does not require high-quality outputs from external models, eliminating the dependency bottleneck~\cite{shao2024deepseekmath, schulman2017proximal}.
Second, RL only requires instructions with reward signals, and constraints in complex instructions are inherently verifiable~\cite{yang2025qwen3}, making RL naturally suited for our framework.
However, applying RL without external supervision presents three key challenges:
(1) Obtaining effective training data and reward signals, as samples that are too hard or too easy can lead to ineffective optimization~\cite{yu2025dapo}.
(2) Modeling rewards for different constraint types, as most works only model hard constraints~\cite{pyatkin2025generalizing}, which can be verified using explicit rules, while many instructions contain soft constraints requiring human judgment~\cite{jiang2023followbench, zhang2024cfbench}.
(3) Training efficiently with RL, as generative reward models are computationally expensive~\cite{qin2025incentivizing}.

To address these challenges, we propose an efficient self-supervised RL framework that improves reasoning models' instruction following capabilities without external supervision.
First, to address sparse learning signals from challenging multi-constraint instructions~\cite{yu2025dapo}, we decompose multi-constraint instructions into simpler instructions with incrementally increasing constraint numbers.
Second, to address soft constraints that require semantic understanding~\cite{ren2025stepbystepmasteryenhancingsoft}, we establish reward signals for soft constraints without any external supervision.
Third, we design an efficient constraint-wise binary classification approach that scores each constraint individually before aggregating the results, achieving computational efficiency while maintaining effectiveness.

Overall, our contributions are as follows:
(1) We propose a self-supervised RL framework that improves instruction following capabilities of reasoning models using only internal signals, eliminating reliance on stronger external models.
(2) We design an efficient reward modeling approach that handles soft constraints in complex instructions through constraint-wise binary classification.
(3) Extensive experiments demonstrate that our framework significantly improves instruction following capabilities while maintaining reasoning performance.

\section{Related Work}
\subsection{Complex Instruction Following Improvement}
Some works distill responses from stronger models through supervised fine-tuning~\cite{sun2024conifer, qin2025incentivizing} or collecting pairwise preference data for direct preference optimization~\cite{he2024complex, qi2024constraint}.
Others adopt self-play approaches that enhance model capabilities through code generation verification~\cite{dong2024self} or training additional refiner models~\cite{cheng2024spar}.
Unlike these approaches, we do not rely on stronger models and instead train exclusively through reinforcement learning based on the model's own capabilities.

\subsection{Reinforcement Learning for Complex Instruction Following}
Some works use rule-based rewards for hard constraints~\cite{lambert2024t, pyatkin2025generalizing}, but these methods cannot generalize to soft constraints.
Other works employ stronger reasoning models~\cite{peng2025verif} or closed-source models~\cite{liu2025recast} as judges for soft constraint reward modeling.
However, they either rely on a stronger model as the reward model or utilize the stronger model to distill data for reward model training.
Additionally, existing approaches often suffer from computational inefficiency.
Different from these works, our reward model is independent of stronger reasoning models, requires no distillation, and operates more efficiently.

\section{Method}

Our framework consists of three main stages. 
First, we construct a multi-constraint instruction dataset and decompose complex instructions into incremental constraint curricula to provide dense training signals. 
Second, as illustrated in Fig.~\ref{fig:framework}, we train constraint-wise binary classification reward models using the self-supervised data, combining them with rule-based verification for comprehensive reward modeling. 
Finally, we apply GRPO algorithm~\cite{shao2024deepseekmath} to optimize the policy model using the composite reward signals.

\subsection{Dataset Construction}

Our training dataset consists of two main components: (1) synthetically generated multi-constraint instructions, and (2) general reasoning data from math and science domains to maintain the model's overall capabilities. 

\paragraph{Complex Instruction Synthesis.}\label{sec:constuct}

To ensure diversity in our multi-constraint instruction dataset, we cover both hard and soft constraint types~\cite{ren2025stepbystepmasteryenhancingsoft}. 
We begin by collecting a diverse set of 3,000 seed instructions from different sources~\cite{kopf2024openassistant,wang2022self,wang2022super,li2025infinity}. Subsequently, we systematically add multiple constraints to these seed instructions, generating comprehensive multi-constraint instruction datasets. For hard constraints, we include 23 types, such as JSON format and frequency of all-capital words. For soft constraints, we include 25 types, such as role-based constraints and emulating specific authors’ styles. Details of the constraint types can be found in Appx.~\ref{appx:ct}. 


\paragraph{General Reasoning Data Integration.}

To maintain the general abilities of the model, we integrate reasoning tasks from math and science domains into our training dataset. Specifically, we select 4,501 math problems from the DeepScaleR-Preview-Dataset~\cite{luo2025deepscaler} and 1,929 science questions from SciKnowEval~\cite{feng2024sciknoweval}. 

\paragraph{Incremental Constraint Curriculum.}\label{sec:const}

Complex instruction-following is challenging~\cite{zhang2024cfbench}, leading to sparse reward signals during RL training~\cite{yu2025dapo}. We decompose complex instructions for progressive learning.
Given a multi-constraint instruction $x$ with constraints $\{c_1 \oplus  c_2 \oplus  \ldots \oplus c_n\}$, we create curriculum levels $\mathcal{L}_k$ where level $k$ contains sub-instruction $x_k$ with the first $k$ constraints:
$
x_k = x \text{ with constraints } \{c_1, c_2, \ldots, c_k\}
$.
This creates a progressive curriculum from single-constraint ($\mathcal{L}_1$) to full multi-constraint instructions ($\mathcal{L}_n$). Statistical details are in Tab.~\ref{tab:num}. 

\newcolumntype{g}{>{\columncolor{green!10}}c}
\setlength\tabcolsep{7pt}
\begin{table}[t]
\centering
\huge
\newcolumntype{b}{>{\columncolor{blue!10}}c}
\renewcommand{\arraystretch}{1.1}
\resizebox{0.4\textwidth}{!}{
\begin{tabular}{cccccc}
\toprule
 Curriculum    & \# Instruct.  &\# Cons.         & \# Soft. & \#Hard.   \\ \midrule
$\mathcal{L}_1$                   &   2806       &2806&1578&1228\\
$\mathcal{L}_2$ &2745&5490 &3203  &2287   \\ 
$\mathcal{L}_3$  &2700     &8100& 4833&3267   \\
$\mathcal{L}_4$          &  2700     &10800   & 6481&4319 \\
$\mathcal{L}_5$              &  2619    &13095 & 8101 &4994   \\ 
\bottomrule
\end{tabular}
}
 \caption{Statistics of curricula. \#Instruct, \#Constraints, \#Soft, and \#Hard refer to the number of instruction constraints, total constraints, soft constraints, and hard constraints, respectively. 
}
 \label{tab:num}
\end{table}
\subsection{Reward Modeling}

To model constraint satisfaction, we design different reward mechanisms for hard and soft constraints to produce constraint-level rewards. 

\paragraph{Hard Constraint Modeling.}
For hard constraints that can be directly verified using explicit rules~\cite{pyatkin2025generalizing}, we adopt programmatic verification. For an input example $(o, c)$, we define a binary constraint-level reward function:
\[
R_{h}(o, c) = 
\begin{cases}
1, & \text{if } o \text{ satisfies constraint }  c \\
0, & \text{otherwise}
\end{cases}
\]


\begin{table}[t]
\centering
\renewcommand{\arraystretch}{1.0}
\resizebox{0.9\columnwidth}{!}{%
\begin{tabular}{lccc}
\toprule
\multicolumn{1}{l}{Reward Model} & \multicolumn{1}{l}{Kendall Tau Coefficient}&\multicolumn{1}{l}{ Position Consistency}\\ \midrule
Our  Dataset                      &  94.0    &  97.0              \\ 
\bottomrule
\end{tabular}
}
\caption{
Agreement Between constructed dataset and human annotation. The detailed evaluation setup is provided in  Appx.~\ref{appx:es}}
  \label{tab:human_label}
\end{table}
\paragraph{Soft Constraint Modeling.}
To model soft constraints that cannot be verified through rules, avoid external supervision, and achieve efficiency, we train a binary classification reward model using self-supervised data from \S\ref{sec:const} without external labels.

During constraint decomposition, a natural relationship emerges: for constraint $c_k$, the response $o_k$ (generated for instruction with constraint $c_k$) is likely to satisfy it, while $o_{k-1}$ (generated for instruction without $c_k$) does not. This allows us to construct training samples: (1) \textit{Positive sample} $(o_k, c_k, label=1)$: response satisfies the constraint, 
(2) \textit{Negative sample} $(o_{k-1}, c_k, label=0)$: response does not satisfy the constraint.
We define a constraint-level reward function $f(o, c) \rightarrow [0, 1]$ that estimates the probability that response $o$ satisfies constraint $c$. The model is trained to minimize the Binary Cross-Entropy loss:

\begin{small}
\[
\mathcal{L} = - \sum_{k=1}^n \left[ \log f(o_k, c_k) + \log \big(1 - f(o_{k-1}, c_k)\big) \right].
\]
\end{small}
As shown in Tab.~\ref{tab:human_label}, our self-supervised dataset demonstrates high consistency with humans. 


\subsection{RL Training}

With the constraint-level reward signals established, we introduce how to utilize these models during reinforcement learning training to predict sample-level rewards for policy optimization.

\paragraph{Reward Model Usage During Training.}
For soft constraints, the trained reward model $f(o, c)$ takes a response $o$ and constraint $c$ as input, producing logits over two classes (satisfy/not satisfy). These logits are converted to probabilities using softmax:

\begin{small}
\[
R_s(o, c) = \frac{\exp(logits[1])}
         {\exp(logits[0]) + \exp(logits[1])}
\]
\end{small}

This gives a scalar reward value $R_s(o, c) \in [0, 1]$ representing the probability that response $o$ satisfies soft constraint $c$. 

\paragraph{Sample-Level Reward Prediction.}
We aggregate constraint-level rewards into sample-level rewards. For instruction $x_k$ with constraints $\{c_1 \oplus c_2, \ldots, c_k\}$ and policy-generated response $o_k$, the sample-level reward is:

\begin{small}
$$
R_f = \frac{1}{k} \sum_{i=1}^{k} r_i, \quad r_i =
\begin{cases}
R_s(o_k, c_i), & \text{if } c_i \text{ is soft} \\
R_h(o_k, c_i), & \text{if } c_i \text{ is hard}
\end{cases}
$$
\end{small}
For reasoning tasks, we assign $R_f = 1$ for correct answers and $R_f = 0$ otherwise.
This composite reward $R_f \in [0, 1]$ captures the overall constraint satisfaction of the response and serves as the reward signal for GRPO optimization. 


\section{Experiment}
\begin{table*}[t]
\centering
\renewcommand{\arraystretch}{1.0}
\resizebox{\textwidth}{!}{
\begin{tabular}{l|c|ccccc|ccc}
\toprule
\multirow{3}{*}{Models} & \multirow{3}{*}{Base Model}
& \multicolumn{5}{c|}{In-Domain} & \multicolumn{3}{c}{Out-of-Domain} \\ \cline{3-7} \cline{8-10}
 & & \multicolumn{1}{c}{IFEval} & \multicolumn{3}{c}{CFBench} & \multicolumn{1}{c|}{FollowBench}  & \multicolumn{1}{c}{ComplexBench}&\multicolumn{1}{c}{WritingBench}& \multicolumn{1}{c}{Collie}  \\ \cline{3-10}
 & & Pr.(L)& CSR & ISR & PSR &HSR & Overall&Avg.&Avg.\\ 
 \midrule
 IR-1.5B$^{*}$ & Qwen-1.5B &57.7 &N/A&N/A&N/A&37.8&N/A&N/A&N/A \\
 Conifer-7B-DPO & Mistral-7B &52.3&55.0&18.0&26.0&50.0&48.1 &3.2&17.8 \\
 Crab-7B-DPO & Mistral-7B &57.7&62.0&25.0&32.0&49.4&59.0 & 4.5&19.6 \\
 TULU 3 Persona IF & LLaMA-8B &69.9&67.0&28.0&38.0&44.2&53.9 &4.4&16.4 \\
 UltraIF-8B-DPO$^{*}$ & LLaMA-8B &75.4&N/A&N/A&N/A&N/A&N/A&N/A&N/A \\
 RECAST-30K-RLVC$^{*}$ & LLaMA-8B &77.4 &N/A&N/A&N/A&\underline{63.2}&N/A&N/A&N/A\\
 SPAR-8B-DPO$^{*}$ & LLaMA-8B &\underline{82.4} &N/A&N/A&N/A&N/A&N/A&N/A&N/A\\
  VERIF-7B$^{*}$ & Qwen-7B &79.5 &N/A&68.0&N/A&N/A&N/A&N/A&N/A\\
 VERIF-8B$^{*}$ & LLaMA-8B &\textbf{87.1} &N/A&\textbf{72.0}&N/A&N/A&N/A&N/A&N/A\\
\midrule
 Qwen2.5-7B-Instruct & Qwen-7B &73.9 &72.0&38.0&48.0 & 55.1& 66.1& 5.7&36.3 \\ 
Qwen2.5-7B-Instruct-R & Qwen-7B &72.3&72.0&36.0&46.0&49.4&63.3&5.8 &35.7\\
 Qwen2.5-7B-Instruct-IF & Qwen-7B &83.9 & 76.0&40.0&52.0 &52.7 &66.7 &5.9&\textbf{38.0} \\

 Distill-Qwen-1.5B-7B & Qwen-1.5B &42.3& 53.0&17.0&24.0&22.7 &39.8 &3.9& 14.0\\ 
 Distill-Qwen-1.5B-IF & Qwen-1.5B &58.8&59.0&20.0&28.0 &26.4 &43.3& 4.1&14.5\\ 
 Distill-Qwen-7B & Qwen-7B &61.7 & 72.0&36.0&48.0 & 41.7&55.2&5.3 &25.2\\ 
 Distill-Qwen-7B-IF &  Qwen-7B &71.7 & 77.0&42.0&52.0 &49.1  &60.7&5.6 &27.0\\ 
 0528-Qwen3-8B & Qwen-8B &79.7 & \underline{89.0}&66.0&\underline{75.0} &60.4&\underline{68.5}&\textbf{7.6}&36.9 \\ 
 0528-Qwen3-8B-IF & Qwen-8B &\textbf{87.1} &\textbf{90.0} &\underline{68.0} &\textbf{76.0} &\textbf{63.8} & \textbf{71.1}&\underline{7.1}&\underline{37.1}\\ 
\bottomrule
\end{tabular}
}
\caption{
The overall performance on In-Domain and  Out-of-Domain benchmarks. We use
  \textbf{bold} for the best results and \underline{underlined} for the second-best results. $^{*}$  denotes results are sourced from the original paper. N/A represents datasets not evaluated in the original papers.
}
\label{tab:main}
\end{table*}

\subsection{Experiment Setup}

We experiment on reasoning LLMs distilled from R1 (R1-Distill-Qwen-1.5B/7B, R1-0528-Qwen3-8B~\cite{guo2025deepseek}). To investigate the effect of incorporating instruction following reasoning data during the cold-start phase of reasoning model training, we also trained a checkpoint Qwen2.5-7B-Instruct-R based on Qwen2.5-7B-Instruct~\cite{team2024qwen2} (details in Appx.~\ref{appx:cs}). \textbf{IF} denotes models trained with our method\footnote{Baselines include instruction-following optimized models detailed in Appx.~\ref{appx:bl}.}.
We evaluate instruction-following ability on multi-constraint benchmarks and general ability on scientific/mathematical reasoning benchmarks\footnote{Details in Appx.~\ref{appx:bec}}.



\subsection{Overall Performance}

As shown in Tab.~\ref{tab:main}, our method effectively improves the model’s instruction-following ability under both hard and soft constraints compared to the backbone models. Moreover, our method is effective across models with different architectures and sizes. Specifically, the performance of the instruct and distilled models shows that our method is applicable to reasoning models with different training backgrounds. 
Detailed results of the model's performance under hard and soft constraints are provided in Appx.~\ref{appx:full}.




\begin{table*}[t]
\renewcommand{\arraystretch}{0.9}
\resizebox{\textwidth}{!}{
\begin{tabular}{lcccccccc}
\toprule

Method & GPQA Diamond  & BBEH & AIME24 & AIME25 & FOLIO & MMLU-Pro&Avg. \\ 
\midrule
Qwen2.5-7B-Instruct  &34.9&44.5&7.2&2.6&57.6
&49.9& 32.8 \\ 
Qwen2.5-7B-Instruct-R&33.8&45.0&8.9&6.0&64.5&51.4&34.9\\
Qwen2.5-7B-Instruct-IF & 34.9&48.5&8.2&5.2&
62.1& 52.7&35.3 \\
 \midrule
R1-Distill-Qwen-7B & 49.1 & 47.5 & 51.2 & 38.7 & 69.5 & 56.1&52.0 \\ 
R1-Distill-Qwen-7B-IF &50.0 & 48.0 & 49.9 & 36.0 & 71.9 & 56.2&52.0 \\ 

\bottomrule
\end{tabular}
}
\caption{Model performance on general capability benchmarks. Avg@30 refers to the average score across 30 sampled responses per question. The max tokens for inference is set to 32k.}
   \label{tab:general}
\end{table*}

\begin{table*}[t]
\centering
\renewcommand{\arraystretch}{1.1}
\resizebox{\textwidth}{!}{
\begin{tabular}{lcccccccc}
\toprule
\multirow{2}{*}{Method} 
& \multicolumn{1}{c}{IFEval} & \multicolumn{3}{c}{CFBench} & \multicolumn{1}{c}{FollowBench} & \multicolumn{1}{c}{ComplexBench} & \multicolumn{1}{c}{WritingBench} & \multicolumn{1}{c}{Collie} \\ 
\cline{2-9}
& Pr. (L) & CSR & ISR & PSR & HSR & Overall & Avg. & Avg. \\ 
\midrule

R1-Distill-Qwen-7B-IF & 71.7 & 77.0&42.0&52.0 &49.1  &60.7&5.6 &27.0 \\ 

\rowcolor[gray]{0.95}
\multicolumn{9}{c}{\textit{Ablation Study}} \\
w/o rule\_based reward & 67.5 &76.0
&40.0&52.0
&  46.7  & 59.0 & 5.6 &20.1   \\ 
w/o probability reward & 69.9 & 77.0 &41.0  &52.0  & 47.1 &58.7  &5.5  &26.9  \\
w/o incremental constraint curriculum & 67.7 & 74.0 &40.0  &49.0  &45.5  & 57.7 & 5.4 &25.1  \\
\bottomrule
\end{tabular}
}
\caption{Ablation study results for  Reward Modeling and Incremental Constraint Curriculum.}
   \label{tab:abla}
\end{table*}
\begin{table}[t]

\renewcommand{\arraystretch}{1.1}
\resizebox{1.0\columnwidth}{!}{%
\begin{tabular}{lccc}
\toprule
\multicolumn{1}{l}{Reward Model} & \multicolumn{1}{l}{Kendall's Tau}&\multicolumn{1}{l}{PC}&\multicolumn{1}{l}{Time Per Group} \\ \midrule
\rowcolor[gray]{0.95}
\multicolumn{4}{c}{\textit{LLM-as-a-judge}} \\
QwQ-32B                      &      73.2      & 87.2                 &  35.7s      \\
IF-Verifier-7B                          &      61.2&82.0  &7.4s    \\ 
\rowcolor[gray]{0.95}
\multicolumn{4}{c}{\textit{Reward Model Training}} \\
BT Training Model                         &48.8 &78.8              &0.4s    \\ 
Our  Reward Model                      &  61.2    &  83.4       &0.3s               \\ 
\bottomrule
\end{tabular}
}
\caption{
Comparison of reward modeling methods.}
  \label{tab:rm}
\end{table}

\begin{figure*}[t] 
    \centering
            \includegraphics[width=1\textwidth]{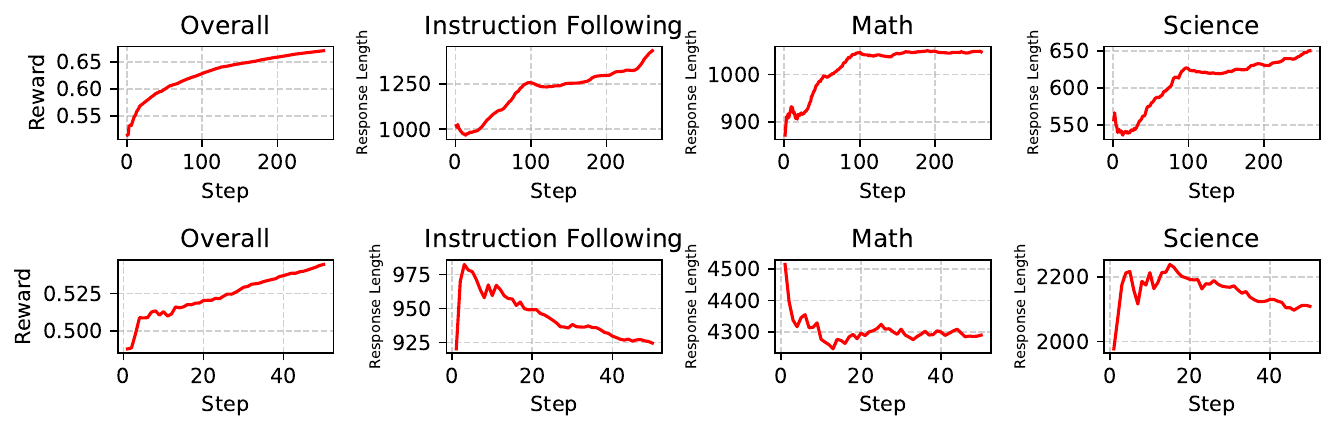}
    \caption{The reward and response length of each domain during RL training of Qwen2.5-7B-R (\textbf{top row}) and R1-Distill-Qwen-7B (\textbf{bottom row}). }
    \label{fig:analyze}
\end{figure*}
\begin{figure}[t] 
    \centering
            \includegraphics[width=0.5\textwidth]{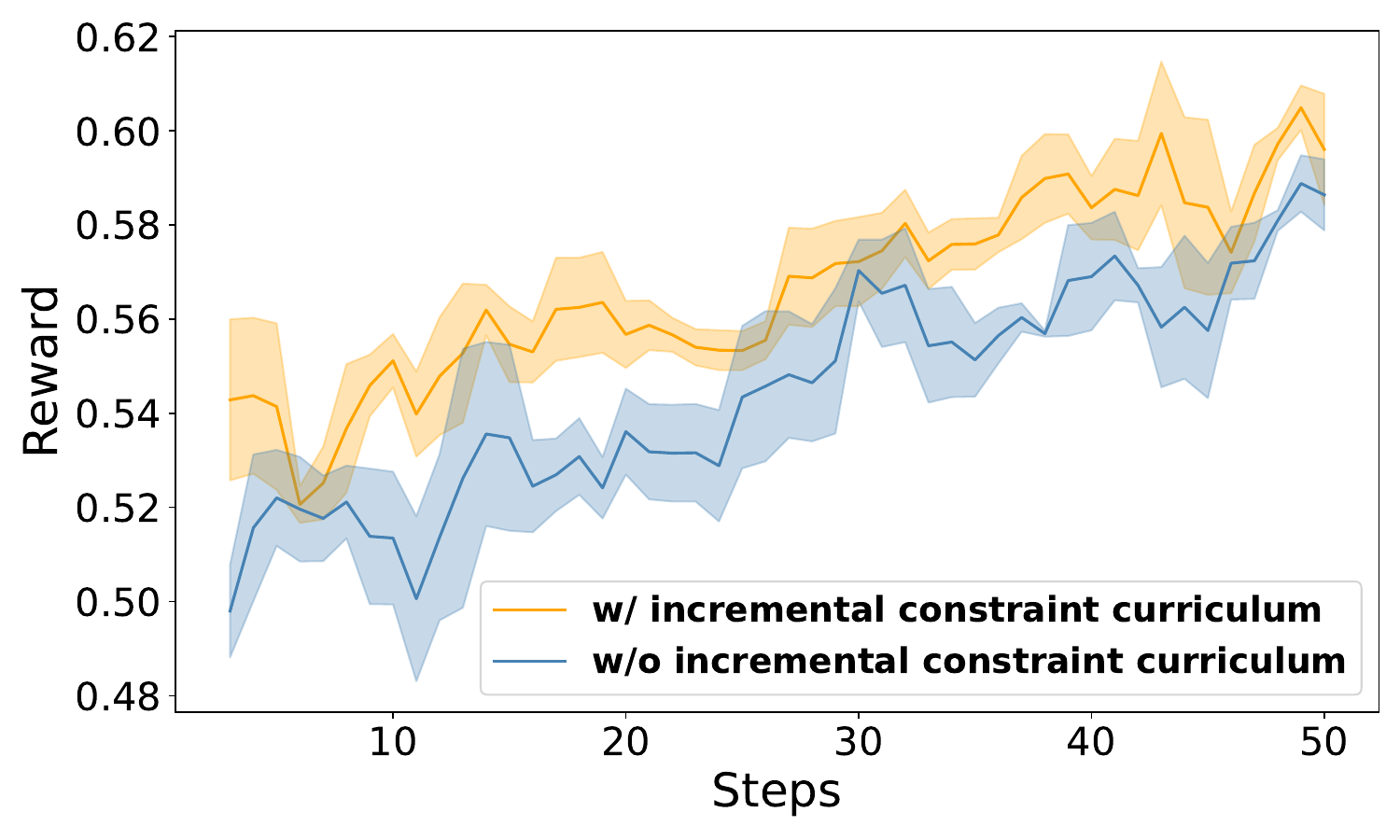}
    \caption{Reward dynamics comparison.}
    \label{fig:com1}
\end{figure}
\subsection{Generalizability}
\paragraph{\textbf{Out-of-Domain Generalization.}}

We select additional benchmarks which contain constraints that are entirely different from training data to evaluate the model's out-of-domain generalization ability. As shown in Tab.~\ref{tab:main}, our method improves the model's instruction-following ability on out-of-domain tasks, demonstrating effective generalization.


\paragraph{\textbf{General Abilities.}}
Besides instruction-following ability, we also evaluate the model's general ability. As shown in Tab.~\ref{tab:general}, our method maintains the model's general abilities while enhancing its instruction-following ability. Moreover, our method can also improve the model's general abilities on some benchmarks, such as FOLIO and MMLU-Pro.

\subsection{Ablation Studies}
\paragraph{\textbf{Reward Modeling.}}
We conduct ablation studies for reward modeling under the following experimental settings: (1) \textbf{w/o rule\_based reward}, which uses only the reward model to provide rewards. (2) \textbf{w/o probability reward}, which directly uses binary rewards: 0 for constraint unfollowing and 1 for constraint following instead of using probability values. As shown in Tab.~\ref{tab:abla}, both settings lead to degraded model performance, indicating the necessity of combining rule-based reward models to prevent reward hacking and probability-based rewards to provide denser supervision. (3) \textbf{Comparison of different modeling methods}. We manually label 50 groups of preference data, each consisting of an instruction with five constraints and five corresponding responses that satisfy 1 to 5 constraints respectively. We then compare the correlation between human-labeled rankings and the rankings produced by three modeling methods: (a) LLM-as-a-judge, (b) the reward model trained with Bradley-Terry (BT) Loss, and (c) our proposed reward model. The LLM-as-a-judge method refers to directly prompting a strong model to assess whether the response satisfies the given constraint, assigning the score of 1 if it does and 0 if it does not. The BT Loss can be defined as:
\begin{equation*}
\mathcal{L} = -\log \sigma(r_a - r_b)
\end{equation*}

Here, $r_a$ denotes the reward of the preferred sample, and $r_b$ denotes the reward of the less preferred sample. As shown in Tab.~\ref{tab:rm}, our reward model demonstrates better alignment with humans and achieves faster inference speed.

\paragraph{\textbf{Incremental Constraint Curriculum.}}
To validate the effectiveness of the Incremental Constraint Curriculum, we conduct ablation studies under the following setting: w/o incremental constraint curriculum, which does not decompose multi-constraint instructions step by step, but instead directly trains on multi-constraint instructions. As shown in Tab.~\ref{tab:abla}, directly training the model with multi-constraint instructions leads to the performance drop. As illustrated in Fig.~\ref{fig:com1}, the model trained w/o incremental constraint curriculum receives sparser rewards during training, which hinders its ability to learn to follow multi-constraint instructions.

\subsection{Training Dynamics}

As shown in Fig.~\ref{fig:analyze}, 
we analyze the dynamics during training on Qwen2.5-7B-R and R1-Distill-Qwen-7B across different task categories.
First, rewards consistently increase during training for both models until convergence. 
Second, response length dynamics reveal critical insights about incorporating reasoning data during cold-start training. 
For Qwen2.5-7B-R, response length increases across all domains, with instruction-following tasks showing the most significant growth. 
This demonstrates that mixing instruction-following reasoning data during cold-start enables the model to generate increasingly longer, more detailed responses.
In contrast, the distilled model shows a different pattern: response length first increases then decreases, with science tasks showing the largest drop. 
Since the distilled reasoning model has been extensively trained on mathematical, programming, and reasoning tasks, its search space has been significantly constrained. 
This finding highlights the importance of incorporating reasoning data during the cold-start phase rather than relying solely on distilled models.

\section{Conclusion}
We propose a self-supervised RL framework that enhances instruction following without external models. 
Our approach features: (1) curriculum decomposition for dense training signals, (2) self-supervised reward modeling using internal capabilities, and (3) constraint-wise binary classification for efficiency.
Experiments show significant improvements in instruction following while preserving reasoning capabilities.

\section{Limitations}
Due to computational resource limitations, we have not validated our method on larger-scale models (e.g., 32B parameters), though our experiments on smaller models provide strong evidence of the method's effectiveness and scalability potential.
Additionally, as our primary focus centers on reward modeling design, the construction of multi-constraint datasets remains relatively limited in diversity, and the current dataset construction process could benefit from incorporating a broader range of constraint types, domains, and complexity levels. 
Despite these limitations, our results provide compelling evidence for the effectiveness of the proposed approach, establishing a solid foundation for future scaling and enhancement efforts.






\bibliography{custom}

\appendix

\section{Appendix}
\subsection{Dataset Analysis}

\begin{figure*}[t] 
    \centering
            \includegraphics[width=1\textwidth]{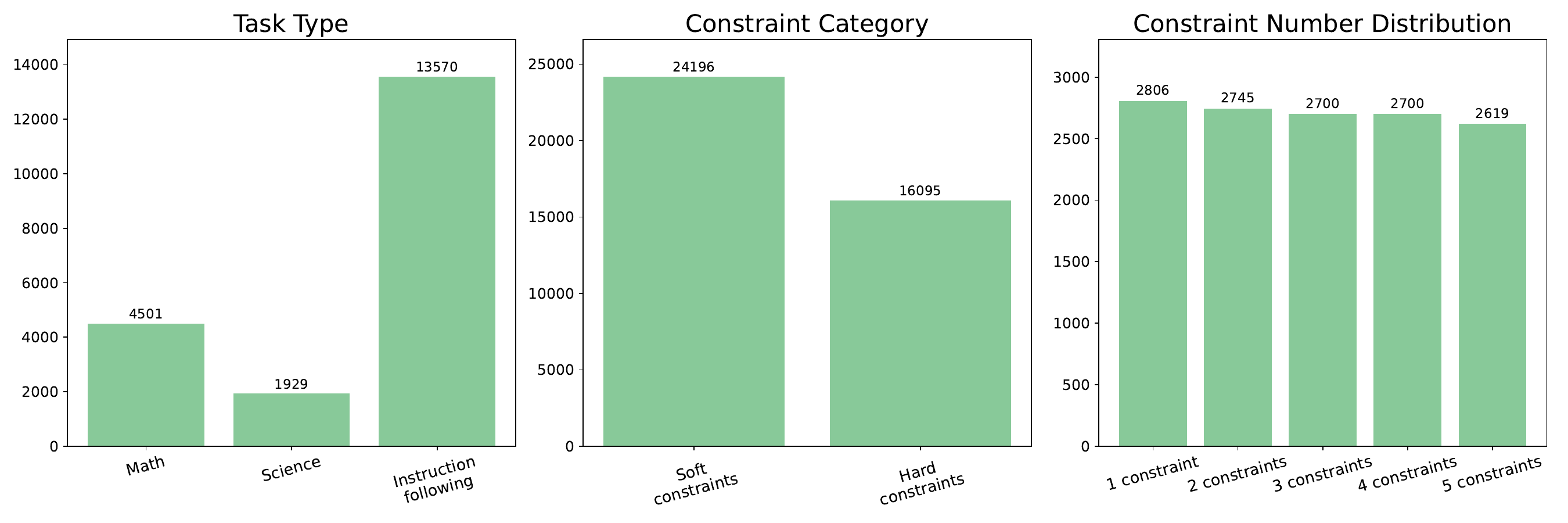}
    \caption{Sample Distribution across Task Types, Constraint Categories, and Number of Constraints}
    \label{fig:dataset}
\end{figure*}
\subsubsection{Constraint Distribution}
Our dataset includes instruction-following, mathematical, and scientific tasks.  For instruction-following tasks, the constraints added in the instructions can be classified into soft constraints and hard constraints. Each instruction contains 1 to 5 constraints. As shown in Fig.~\ref{fig:dataset}, we present statistics on the amount of data across different domains, the number of soft and hard constraints, and the distribution of instructions with varying numbers of constraints. As shown in Tab.~\ref{tab:app_soft} and Tab.~\ref{tab:app_hard}, we present examples of soft and hard constraints in the dataset. 

\subsubsection{Constraint Types}
\label{appx:ct}
As shown in Tab.~\ref{tab:constraint-types}, our dataset includes various types of constraints, covering multiple domains and tasks. For each seed instruction, we use the prompt shown in Tab.~\ref{tab:prompt-raw} with GPT-4o to construct constraints, resulting in multi-constraint instructions.

\begin{table*}
\resizebox{\linewidth}{!}{
\begin{tcolorbox}
\small

\textit{\color{gray}{/* Seed instruction */}} \\
Assign a category to each text snippet identifying the main emotion expressed. Categories may include ``joy,'' ``anger,'' ``sadness,'' or ``fear.''  Sentence: NAME\_2 was overwhelmed with tears of happiness as she accepted the award on stage.

\textit{\color{gray}{/* Constraints */}} \\
Must include the word ``exuberant'' in at least one category description.

Incorporate a scenario where a character is experiencing mixed emotions.

Limit the response to a maximum of 25 words.

Write the response in the style of a psychological evaluation report.

Tailor the answer for an audience of high school psychology students.

\end{tcolorbox}
}
\caption{Examples of instructions with soft constraints.}
\label{tab:app_soft}
\end{table*}
\begin{table*}
\resizebox{\linewidth}{!}{
\begin{tcolorbox}
\small

\textit{\color{gray}{/* Seed instruction */}} \\
Given a question, generate a paraphrase of that question without changing the meaning of it. Your answer should reword the given sentence, but not add information to it or remove information from it. The answer to your question should be the same as the answer to the original question. Input: Question: who does the voice of carl in phineas and ferb?

\textit{\color{gray}{/* Constraints */}} \\
Answer with the letter \texttt{"i"} appearing at least 4 times. 

Highlight at least 2 sections of your response in markdown such as \texttt{*highlighted section*}.

Make sure your reply is in English and all capital letters. 

Put your entire response inside double quotation marks.

\end{tcolorbox}
}
\caption{Examples of instructions with hard constraints.}
\label{tab:app_hard}
\end{table*}
\begin{table*}
\resizebox{\linewidth}{!}{
\begin{tcolorbox}
\small

\textit{\color{gray}{/* Positive Example */}} \\
\textbf{Question:} Introduce Time Constraints: Impose a time limit for the role-play to unfold or for certain actions to be completed. \\
\textbf{Answer:} Characters identified in the story: 1. Kathy 2. Sue. The main conflict between the characters is the tension and argument that arises due to Kathy spreading rumors about Sue and Sue discovering and reacting to these rumors. To introduce a time constraint for the role-play, we can say: The events in the story unfold over the course of a week. Kathy starts the rumors on Monday, Sue finds out by Wednesday, they have the argument on Thursday, and by the end of the week, Sue stops speaking to Kathy. This time constraint adds a sense of urgency and progression to the narrative. \\
\textbf{Label:} 1

\textit{\color{gray}{/* Negative Example */}} \\
\textbf{Question:} Introduce Time Constraints: Impose a time limit for the role-play to unfold or for certain actions to be completed. \\
\textbf{Answer:} The characters available in the given story are Kathy and Sue. The main conflict between the characters is the tension and argument that arises due to Kathy's rumors about Sue, which leads to Sue finding out and subsequently never talking to Kathy again. \\
\textbf{Label:} 0

\end{tcolorbox}
}
\caption{Examples of reward model training data.}
\label{tab:app_rm}
\end{table*}

\begin{table*}
\centering
\renewcommand{\arraystretch}{1.5}
\small
\begin{tabular}{|p{3.5cm}|p{4.8cm}|p{5.5cm}|}
\hline
\textbf{Constraint Type} & \textbf{Definition} & \textbf{Example} \\
\hline
Lexical content constraint & Must include specific terms or symbols with precise placement. & "...must include the word 'beautiful.'" \\
\hline
Element constraint & Include specific entities or scenarios. & "...highlights the Great Wall." \\
\hline
Semantic constraint & Focus on themes, tone, or stance. & "Write a poem about London." \\
\hline
Word Count & Limit the number of words. & "A 50-word poem." \\
\hline
Sentence Count & Limit the number of sentences. & "...three sentences." \\
\hline
Paragraph Count & Limit the number of paragraphs. & "...divided into 3 sections." \\
\hline
Document Count & Limit the number of documents. & "...list 3 articles." \\
\hline
Tone and emotion & Conform to specific emotional tone. & "Write a letter in an angry and sarcastic tone." \\
\hline
Form and style & Use specified stylistic form and perception. & "Write a passage in an encyclopedic style." \\
\hline
Audience-specific & Tailored to a specific audience group. & "Write a poem for a 6-year-old." \\
\hline
Authorial style & Emulate specific authors’ styles. & "Write a passage in the style of Shakespeare." \\
\hline
Fundamental format & Follow standard formats like JSON, HTML, etc. & "Output in JSON format." \\
\hline
Bespoke format & Use custom formatting protocols. & "Bold the main idea and output in unordered list." \\
\hline
Specialized format & Tailored for specific applications or domains. & "Convert to electronic medical record format." \\
\hline
Pragmatic constraint & Adapt to context like dialects or language policy. & "Output in English, classical Chinese, etc." \\
\hline
Syntactic constraint & Follow specific phrase and clause structures. & "Use imperatives with nouns and verb phrases." \\
\hline
Morphological constraint & Control over affixes, roots, and word formation. & "Output all content in lowercase English." \\
\hline
Phonological constraint & Focus on sounds, tone, and intonation. & "Single-syllable tongue twisters." \\
\hline
Role-based constraint & Respond with specific role identity. & "You are Confucius, how do you decide?" \\
\hline
Task-specific constraint & Address a defined situational task. & "Work from home, how to report?" \\
\hline
Complex context constraint & Involve multi-faceted and nested reasoning. & "On the left, 10 total, what to do?" \\
\hline
Example constraint & Conform to patterns from example pairs. & "input:x..., output:{...}; input:y..., output?" \\
\hline
Inverse constraint & Narrow response space via exclusions. & "No responses about political topics." \\
\hline
Contradictory constraint & Combine requirements that are hard to satisfy simultaneously. & "A five-character quotation, 1000 words." \\
\hline
Rule constraint & Follow symbolic or logical operation rules. & "Each answer adds 1+1=3, then 2+2=5." \\
\hline
\end{tabular}
\caption{Constraint Types~\cite{zhang2024cfbench}.}
\label{tab:constraint-types}
\end{table*}

\begin{table*}
\centering
\renewcommand{\arraystretch}{1.2}
\small
\begin{tabular}{|p{15cm}|}
\hline
[Task Description] \\
1. I currently have a seed question, but the seed questions are relatively simple. To make the instructions more complex, I want you to identify and return five atomic constraints that can be added to the seed question. \\
2. I will provide [Seed Question] and [Constraint References], and you can use these references to propose five constraints that would increase the difficulty of the seed question. \\
3. [Constraint References] are just suggestions. You may choose one or more constraints from the list or propose new ones if needed. \\
4. Do not modify or rewrite the seed question. Your task is only to generate new constraints that can be added to it. \\
5. Return the added constraints in the following JSON format: \\
\quad json \\
\quad \{ \\
\quad\quad "c1": "<first constraint>", \\
\quad\quad "c2": "<second constraint>", \\
\quad\quad "c3": "<third constraint>", \\
\quad\quad "c4": "<fourth constraint>", \\
\quad\quad "c5": "<fifth constraint>" \\
\quad \} \\
6. Do not return anything else. No explanation, no reformulated question, no analysis---only the JSON structure. \\
\\

[Constraint References] \\
1. Lexical content constraint : \{Definition\} \{Example\} \\
2. Word Count : \{Definition\} \{Example\} \\
$\ldots$ \\
25. Rule Constraint : \{Definition\} \{Example\} \\
\\

[Seed Question] \\
\{raw\_question\} \\
\hline

\end{tabular}
\caption{Prompt for generating constraints.}
\label{tab:prompt-raw}
\end{table*}

\begin{table*}
\resizebox{\linewidth}{!}{
\begin{tcolorbox}
\small

You are a meticulous assistant who precisely adheres to all explicit and implicit constraints in user instructions. When presented with complex, multi-constraint tasks, you follow a structured approach:\\

FORMAT: Your response must be structured with two distinct sections:\\
<think> Your detailed analytical process and strategy development </think>\\
<answer> Your final solution that adheres to all requirements </answer>\\

In your <think> section, employ these strategic approaches:\\
1. Constraint Analysis: Thoroughly examine each explicit and implicit constraint, identifying their interdependencies and potential conflicts.\\
2. Validation Strategy: Draft preliminary answers, then methodically verify compliance with each constraint, making iterative refinements as needed.\\
3. Sequencing Logic: Determine the optimal order for addressing constraints, prioritizing foundational requirements before tackling dependent ones.\\
4. Edge Case Consideration: Proactively identify boundary conditions and exceptions that might challenge constraint compliance.\\
5. Coherence Check: Ensure your solution maintains internal consistency while satisfying all requirements simultaneously.\\
6. Format Verification: Confirm your response adheres to all specified formatting and structural guidelines.\\

In your <answer> section, deliver a solution that precisely implements all requirements while maintaining natural flow and coherence. Your final response must satisfy all constraints without drawing attention to the mechanics of constraint management itself.\\

Here is an example: 

\{example\}
 
\end{tcolorbox}
}
\caption{Prompt used to construct data for cold start.}
\label{tab:app_cold}
\end{table*}

\subsubsection{Evaluation Setup}
\label{appx:es}
We manually label 50 groups of preference data, each consisting of an instruction with five constraints and five corresponding responses that satisfy 1 to 5 constraints respectively. We then compare the correlation between the human-labeled ranking and the ranking.

\subsection{Reward Model Training}

We perform full fine-tuning on two models, Qwen2.5-1.5B-Instruct and Qwen2.5-7B-Instruct, for a binary classification task aimed at determining whether a response satisfies a given constraint. Each training example consists of a prompt that concatenates the response and its associated constraint. Fine-tuning is conducted using the HuggingFace Trainer framework with FP16 precision enabled and Deepspeed optimization (Stage 2 ZeRO) configured via a JSON file specifying automatic batch size and gradient accumulation steps, as well as support for the adamw\_torch optimizer. Training uses a batch size of 1, a learning rate of 5e-6, and runs for 3 epochs. Accuracy is employed as the evaluation metric. The Qwen2.5-1.5B reward model provides reward signals for reinforcement learning (RL) training of Distill-Qwen-1.5B, while the Qwen2.5-7B reward model serves as a general reward function for other base models. All training is performed on 8 NVIDIA A100 80GB GPUs. As shown in Tab.~\ref{tab:app_rm}, we present examples of the reward model training data.

\subsection{RL Training}
\subsubsection{Implementation Details}
We apply the GRPO training using the VeRL framework. We use a distributed training setup across 3 nodes, each equipped with 8 NVIDIA A100 80GB GPUs, for a total of 24 NVIDIA A100 80GB GPUs. Prompts and responses are truncated to a maximum length of 8192 tokens.
The RL algorithm employs the GRPO advantage estimator with no KL penalty (kl\_coef set to 0.0). The optimizer uses a learning rate of 1e-6 with a weight decay of 1e-2, and no learning rate warm-up. We leverage Fully Sharded Data Parallel (FSDP) with full sharding enabled, rank0 parameter initialization, and parameter offloading to optimize GPU memory usage. The training batches are organized with a global batch size of 96, micro-batches of size 2 for updates, and micro-batches of size 16 for experience generation. Gradient clipping is applied with a max norm of 1.0. Rollouts are performed with a temperature of 1.0 and a group size of 5. Tensor parallelism of size 2 is applied. We train for a total of 260 steps.

\subsubsection{Baselines}
\label{appx:bl}
We include IR-1.5B~\cite{qin2025incentivizing}, Conifer-7B-DPO~\cite{sun2024conifer},
Crab-7B-DPO~\cite{qi2024constraint}, TULU 3 Persona IF~\cite{lambert2024t}, UltraIF-8B-DPO~\cite{an2025ultraif}, RECAST-30K-RLVC~\cite{liu2025recast}, SPAR-8B-DPO~\cite{cheng2024spar}, and VERIF~\cite{peng2025verif} as our baselines.

\subsubsection{Benchmarks}
\label{appx:bec}
We evaluate instruction-following ability using benchmarks such as IFEval~\cite{zhou2023instruction}, CFBench~\cite{zhang2024cfbench}, FollowBench~\cite{jiang2023followbench}, ComplexBench~\cite{wen2024benchmarking}, WritingBench~\cite{wu2025writingbench}, and Collie~\cite{yao2023collie}.
In contrast, we assess general reasoning and knowledge capabilities with datasets including GPQA-Diamond~\cite{rein2024gpqa}, BBEH~\cite{kazemi2025big}, AIME2024~\cite{AIME2024}, AIME2025~\cite{AIME2025}, FOLIO~\cite{han2022folio}, and MMLU-Pro~\cite{wang2024mmlu}.

\subsubsection{Cold Start}
\label{appx:cs}
Effective instruction following requires systematic thinking patterns that can handle complex multi-constraint scenarios. We identify that thinking patterns guided by Constraint Satisfaction Problems (CSP) are particularly beneficial for instruction following, including: (1) \textbf{Constraint Decomposition} - breaking down complex instructions into individual constraint components; (2) \textbf{Constraint Prioritization} - identifying and ordering constraints by importance and feasibility; (3) \textbf{Incremental Consistency Checking} - verifying constraint satisfaction step-by-step during response generation; and (4) \textbf{Backtracking Search} - revising responses when constraint conflicts are detected. To incorporate these CSP-guided thinking patterns, we design a specialized prompt that encourages models to explicitly reason through constraint satisfaction processes. We construct 1,361 instructions containing 1 to 7 constraints, generate responses using Claude-Sonnet-3.7-thinking as instruction following reasoning data, and then use these data for supervised fine-tuning (SFT) training. The prompt we use is detailed in Tab.~\ref{tab:app_cold}.

We fine-tune the Qwen2.5-7B-Instruct model using full-parameter training with DeepSpeed ZeRO-3 optimization using LLaMA-Factory~\cite{zheng2024llamafactory} on three  NVIDIA A100 80GB GPUs. The training runs for 2 epochs with a batch size of 2 per device and a gradient accumulation step of 2. We set the learning rate to 1e-5 with a cosine scheduler and a warmup ratio of 10\%. The training data is processed using 16 preprocessing workers, with a maximum sequence length of 4096 tokens. Mixed-precision training is enabled using bfloat16 (bf16).

\subsubsection{Analysis of Model Response Improvement}

Tab.~ \ref{tab:cat_poem_comparison} demonstrates the model’s improvement after training. Before training, the model mainly generates a detailed thought process instead of directly producing the poem, resulting in an output that does not meet the specified constraints and has a low score. After training, the model outputs a concise poem that strictly follows all requirements—including the title format, sentence limit, and punctuation rules—with fluent and vivid content. This clear difference in output quality and adherence to constraints, along with the improved score, shows that the training effectively enhances the model’s performance.

\begin{table*}
\centering
\renewcommand{\arraystretch}{1.2}
\small
\begin{tabular}{|p{0.2\linewidth}|p{0.82\linewidth}|}
\hline
\textbf{Case} & \textbf{Content} \\
\hline

\textbf{Before Training} & 
\textbf{Input:} Write a poem about a curious cat. The poem must have a title wrapped in double angular brackets, i.e. <<title>>, contain less than 13 sentences, and no commas. Don't forget to add other punctuations.<think>

\textbf{Think:} Alright, so the user has asked me to write a poem about a curious cat. They've given me specific guidelines: the title needs to be in double angular brackets, the poem should have less than 13 sentences, no commas, and include other punctuations. Hmm, okay, let me break this down.

First, I need a title. It should be catchy and reflect the curiosity aspect. Maybe something like "Whisker's Quest" or "Paws of Curiosity." I'll go with "Curious Cat's Journey" since it clearly states the theme.

Next, the structure. The poem should have fewer than 13 sentences, so maybe around 11 to keep it concise. I'll aim for a consistent rhyme scheme to make it flow smoothly. Let's try AABB or ABAB, but I'll see as I go.

No commas, so I'll have to be careful with sentence structure. Instead, I'll use other punctuation like periods, exclamation points, and question marks to add rhythm and emphasis. Maybe some internal rhymes to make it catchy.

Thinking about the content, I want to highlight the cat's curiosity and adventure. Maybe describe the cat exploring places, meeting new things, and the excitement of discovery. Including sensory details like sights, sounds, smells can make the poem vivid.

I'll start each line with a different image related to curiosity—like whiskers, paws, exploring, meeting new friends, etc. Then, gradually build up the story of the cat's journey. Each stanza can focus on a different aspect: meeting a friend, discovering a new place, the thrill of exploration.

I should also include a question to engage the reader, something like "What's next for our curious friend?" to leave a lasting impression. Ending with a playful line about the cat chasing its own shadow adds a nice touch and a bit of whimsy.

Now, putting it all together, I'll draft each line, ensuring it flows without commas and uses the required punctuation effectively. I'll count the sentences to make sure it's under 13. Let me check each line for rhythm and how well it contributes to the overall theme of curiosity and adventure.

Finally, I'll review the poem to ensure it meets all the user's guidelines and conveys the intended emotions and imagery. Making sure it's engaging and captures the essence of a curious cat's journey.
</think>

\textbf{Output:} Model responds with a detailed plan and thought process instead of directly generating the poem. It includes reasoning about how to choose the title, avoid commas, structure the stanzas, and ensure vivid imagery. The final poem is long, significantly exceeding the 13-sentence limit.

\textbf{Score:} 0.333 \\

\hline

\textbf{After Training} & 
\textbf{Input:} Write a poem about a curious cat. The poem must have a title wrapped in double angular brackets, i.e. <<title>>, contain less than 13 sentences, and no commas. Don't forget to add other punctuations.<think>

\textbf{Think:} Alright, let's break down the user's query. They want a poem about a curious cat with specific constraints. The title needs to be in double angular brackets, less than 13 sentences, no commas, and include other punctuations.

First, I'll think about the structure. The poem should be concise, so each line should pack a punch. I'll start with a title that reflects curiosity, maybe something like "The Curious Cat's Tail" to fit the format.

Next, I need to ensure each line flows well without commas. I'll use periods or other punctuation like exclamation marks or question marks to maintain rhythm. I'll describe the cat's actions and environment to make it vivid.

I'll consider the imagery. A curious cat might explore, interact with other animals, and show personality. Including elements like yarn, yarn balls, and sun can add depth. I'll make sure each line is short and impactful.

Now, checking the sentence count. I'll aim for around 10-12 lines to stay under 13. I'll also ensure that the poem doesn't use any commas, so I'll structure the sentences to avoid them naturally.

Finally, I'll review the poem to make sure it meets all the criteria: title format, sentence count, no commas, and appropriate punctuation. This should create a engaging and concise poem that highlights the cat's curiosity.
</think>

\textbf{Output:}

<<The Curious Cat's Tail>> A curious cat purred soft and high  
Curious as a yarn ball on a hook  
It meandered through the sunlit yard  
Exploring every nook and every pore  
Meeting friends old and new along the way  
With curiosity in its bright eyes  
It spun tales of adventure untold  
Of yarn and games and endless joy  
The cat's tail wagged like a flag  
Pointing to wonders both near and far  
In each glance and in each sigh  
The curious cat's spirit did thrive  
A poem of paws and a heart full of glee  
Wrapped in warmth and filled with glee

\textbf{Score:} 1.0 \\
\hline

\end{tabular}
\caption{Comparison of the outputs of DeepSeek-R1-Distill-Qwen-7B before and after training for the same instruction.}
\label{tab:cat_poem_comparison}
\end{table*}

\begin{figure*}[t] 
    \centering
            \includegraphics[width=1\textwidth]{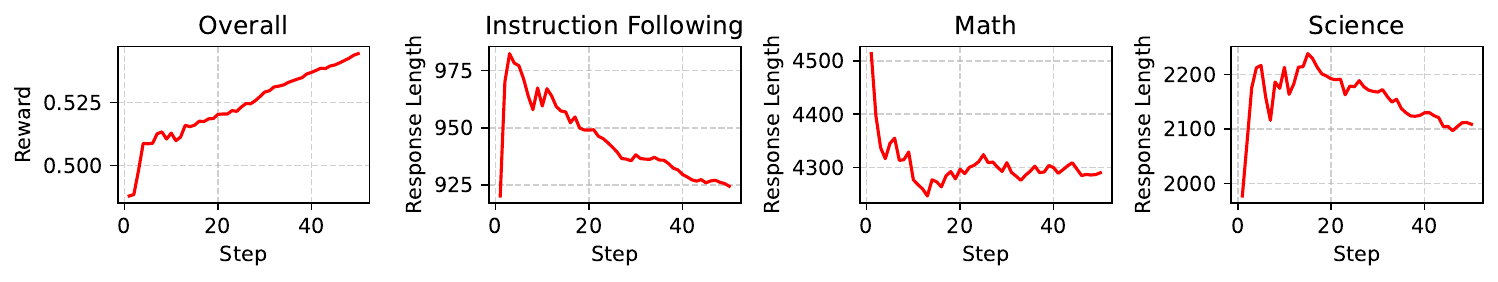}
    \caption{The reward and response length of w/o incremental constraint curriculum. }
    \label{fig:analyze1}
\end{figure*}

\subsubsection{Full Training Dynamics Analysis}
As shown in Fig.~\ref{fig:analyze1}, we present the model’s reward and output length changes during training under the setting without incremental constraint curriculum (w/o incremental constraint curriculum).

\subsubsection{Full Performance on Soft and Hard Constraints} \label{appx:full}
As shown in Tab.~\ref{tab:soft_cf} and Tab.~\ref{tab:hard}, we present the model’s performance on the soft and hard constraint benchmarks.
\begin{table*}[t]
\newcolumntype{g}{>{\columncolor{green!10}}c}
\newcolumntype{b}{>{\columncolor{blue!10}}c}
\renewcommand{\arraystretch}{0.9}
\renewcommand{\familydefault}{\rmdefault}
\resizebox{\textwidth}{!}{
\begin{tabular}{lccccccccc}
\toprule
\multirow{2}{*}{Method} & \multicolumn{3}{c}{Easy Set} & \multicolumn{3}{c}{Hard Set} & \multicolumn{3}{c}{Full Set}  \\
\cmidrule(lr){2-4} \cmidrule(lr){5-7} \cmidrule(lr){8-10}
& CSR & ISR & PSR & CSR & ISR & PSR & CSR & ISR & PSR  \\
\midrule

Qwen2.5-7B-Instruct &79.0 & 52.0 & 58.0 & 66.0 & 24.0 & 37.0 & 72.0 & 38.0 & 48.0 \\
Qwen2.5-7B-Instruct-R &80.0 & 52.0 & 58.0 & 65.0 & 20.0 & 34.0 & 72.0 & 36.0 & 46.0\\

Qwen2.5-7B-Instruct-IF&81.0 & 55.0 & 63.0 & 71.0 & 24.0 & 41.0 & 76.0 & 40.0 & 52.0 \\
\midrule
R1-Distill-Qwen-1.5B &58.0 & 24.0 & 29.0 & 48.0 & 10.0 & 18.0 & 53.0 & 17.0 & 24.0 \\
R1-Distill-Qwen-1.5B-IF & 64.0 & 30.0 & 35.0 & 53.0 & 11.0 & 20.0 & 59.0 & 20.0 & 28.0 \\
R1-Distill-Qwen-7B &79.0 & 50.0 & 57.0 & 66.0 & 22.0 & 38.0 & 72.0 & 36.0 & 48.0 \\
R1-Distill-Qwen-7B-IF & 83.0 & 56.0 & 62.0 & 71.0 & 27.0 & 41.0 & 77.0 & 42.0 & 52.0 \\
R1-0528-Qwen3-8B &95.0 & 83.0 & 86.0 & 84.0 & 50.0 & 64.0 & 89.0 & 66.0 & 75.0 \\  
R1-0528-Qwen3-8B-IF &94.0 & 83.0 & 86.0 & 85.0 & 54.0 & 66.0 & 90.0 & 68.0 & 76.0 \\  
\bottomrule
\end{tabular}
}
\caption{
Full results on CFBench. CSR, ISR, and PSR refer to Constraint Satisfaction Rate, Instruction Satisfaction Rate, and Priority Satisfaction Rate, respectively.
}
\label{tab:soft_cf}
\end{table*}
\begin{table}[t]

\renewcommand{\arraystretch}{0.9}
\resizebox{1.0\columnwidth}{!}{
\begin{tabular}{lccccc}
\toprule
Method & Pr. (S) & Ins. (S) & Pr. (L) & Ins. (L) & Avg. \\ 
\midrule

Qwen2.5-7B-Instruct & 71.5 & 79.1 & 73.9 & 81.3 & 76.5 \\
Qwen2.5-7B-Instruct-R & 71.9
&78.7&72.3&79.0&75.5 \\
Qwen2.5-7B-Instruct-IF & 83.7 & 88.7 & 83.9 & 88.8 & 86.3 \\

\midrule

R1-Distill-Qwen-1.5B & 38.1 & 50.4 & 42.3 & 54.3 & 46.3\\ 
R1-Distill-Qwen-1.5B-IF & 54.0 & 65.8 & 58.8 & 69.8 & 62.1 \\
R1-Distill-Qwen-7B & 56.7 & 68.3 & 61.7 & 72.5 & 64.8 \\ 
R1-Distill-Qwen-7B-IF & 65.8 & 75.2 & 71.7 & 80.2 & 73.2 \\ 
R1-0528-Qwen3-8B &74.9 & 82.4 & 79.7 & 85.9 & 80.7\\ 
R1-0528-Qwen3-8B-IF &83.5 & 88.7 & 87.1 & 91.2 & 87.6 \\
\bottomrule
\end{tabular}
}
\caption{Full results on IFEval. “Pr.” and “Ins.” refer to prompt-level and instruction-level metrics, respectively, while “S” and “L” denote strict and loose evaluation.}
   \label{tab:hard}
\end{table}

\end{document}